\documentclass[letterpaper, 10 pt, conference]{ieeeconf}
\IEEEoverridecommandlockouts
\usepackage[usenames,dvipsnames]{color} 
\usepackage{amsmath, amsfonts, bm}
\usepackage{graphicx}
\usepackage{tabularx}
\newcolumntype{C}[1]{>{\centering\arraybackslash}p{#1}}

\usepackage{booktabs}
\usepackage{mathtools}
\usepackage{comment}
\usepackage{textgreek}
\usepackage{hyperref}
\usepackage{nicefrac, xfrac}
\usepackage{subcaption}

\newcommand{\distanceGaussian}{\bm{D}}

\newcommand{\distanceWieghtedAvg}{\bm{W}}

\newcommand{\shape}{\bm{I}}

\newcommand{\numberOfPoints}{n}

\title{\emph{Sim2real} Cattle Joint Estimation in 3D point clouds}
\author{   
    Mohammad Okour, Raphael Falque and Alen Alempijevic\\
    Robotics Institute, University of Technology Sydney, NSW, Australia \\
    \{mohammad.okour\}@student.uts.edu.au,\\
    \{raphael.falque, alen.alempijevic\}@uts.edu.au
}

\begin{document}

\maketitle

\begin{abstract}
    Understanding the well-being of cattle is crucial in various agricultural contexts. Cattle's body shape and joint articulation carry significant information about their welfare, yet acquiring comprehensive datasets for 3D body pose estimation presents a formidable challenge. This study delves into the construction of such a dataset specifically tailored for cattle. Leveraging the expertise of digital artists, we use a single animated 3D model to represent diverse cattle postures. To address the disparity between virtual and real-world data, we augment the 3D model's shape to encompass a range of potential body appearances, thereby narrowing the "sim2real" gap. We use these annotated models to train a deep-learning framework capable of estimating internal joints solely based on external surface curvature. Our contribution is specifically the use of geodesic distance over the surface manifold, coupled with multilateration to extract joints in a semantic keypoint detection encoder-decoder architecture. We demonstrate the robustness of joint extraction by comparing the link lengths extracted on real cattle mobbing and walking within a race. Furthermore, inspired by the established allometric relationship between bone length and the overall height of mammals, we utilise the estimated joints to predict hip height within a real cattle dataset, extending the utility of our approach to offer insights into improving cattle monitoring practices.
\end{abstract}

\section{Introduction}
In modern agriculture, robotics and automation promise a transformative future, streamlining labour-intensive tasks and enhancing productivity \cite{duong2020review}. Perception systems are critical for livestock production systems, the animal body structure has an impact on behavior, well-being, and fertility \cite{saad2021foot,b2017association}. Specifically, observing cattle during locomotion is crucial for identifying health issues in livestock, such as structural soundness and lameness \cite{b2017association,bell2018,russello2022t}. Crucial in this assessment is identifying the pose of joints and limb actuation, allowing for body pose estimation. Advancements in perception offer invaluable insights for optimising livestock management practices and improving overall herd performance.

Human pose detection and tracking frameworks have garnered significant attention for their versatile applications in human-computer interaction and activity recognition \cite{wang2021deep}. However, the scarcity of annotated animal pose data presents a major obstacle to developing animal pose estimation approaches. Animals, unlike humans, lack the capability to cooperate during data collection, resulting in significant difficulty in coordinating them in the process. Available animal data sets lack ground truth for joint position and solely contain human-annotation \cite{yu2021ap, jianguo2019iccv, joska2021acinoset}. 

Synthetic training data for 3D animal pose estimation has been explored, adapting techniques from human pose estimation \cite{zoobuilder2020}. This approach generates RGB images for network training, focusing on minimising distributional differences between synthetic and real animal data. However, the approach generally considers isolated animals rather than herds and does not fully exploit joint and shape information for comprehensive animal assessment. Moreover, existing simulations lack intra-species variability \cite{zoobuilder2020}, limiting their ability to address \emph{sim2real} challenges, particularly regarding shape deformation during animal movement \cite{hofer2021sim2real}. Our previous work \cite{Falque2022,okour2022sim2real} captures from multi-depth cameras capture 3D data of cattle in high fidelity while they are travelling through a race.

\begin{figure}
  \centering
  \includegraphics[width=0.49\linewidth]{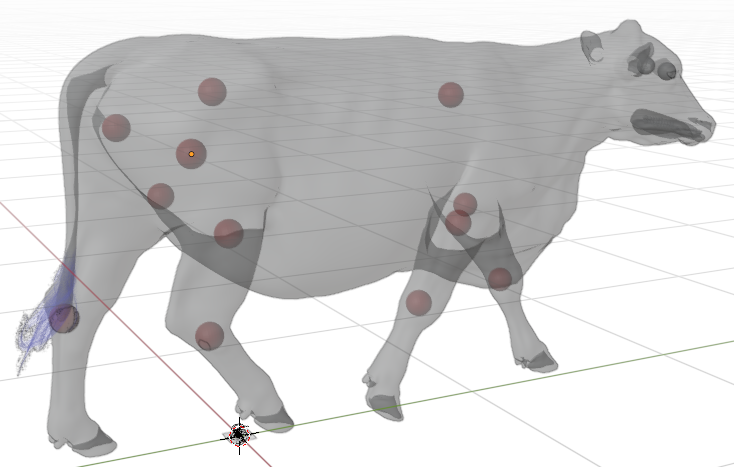}
  \includegraphics[width=0.49\linewidth]{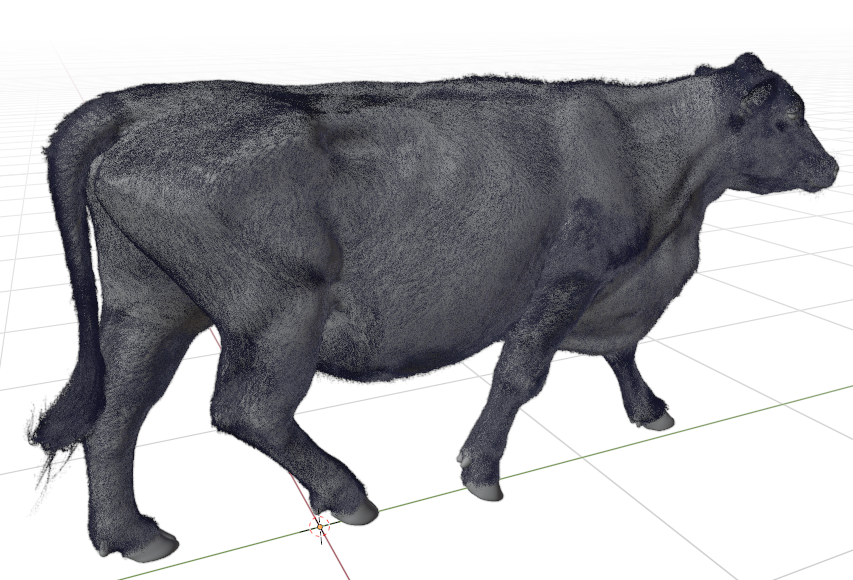}
  \caption{Each model is annotated by twelve joints: two joints at each of the front legs, two at each of the back legs, two at each side of the hip bones, and two at either end of the spine.}
  \label{fig:Annotated model}
\end{figure}

Extending our work on estimating joint coordinates from 3D point cloud data  \cite{okour2022sim2real}, we propose a methodology to utilise the manifold defined by the surface while leveraging animated 3D models to represent diverse cattle postures. Our contribution is specifically the use of geodesic distance over the surface manifold, coupled with multilateration to extract joints in a semantic keypoint detection encoder-decoder architecture. This enables the extraction of joint locations,  where we demonstrate the robustness of our method to estimate joint locations on real cattle while they are in motion. We utilise joint information to estimate the hip height of cattle, drawing from work \cite{ali1984relationship} that demonstrates an allometric relationship. 

\section{Related Work}
Pose estimation plays a crucial role across various disciplines, whether for humans or animals. In human-centric applications, such as virtual reality, gaming consoles, and human activity recognition, accurate pose estimation is paramount. Early research by Shotton et al. \cite{shotton2011} demonstrated the potential of depth data for human pose estimation, primarily in controlled environments like motion capture rooms. Subsequently, RGB-based approaches have gained momentum, with researchers like Pavlakos et al. \cite{pavlakos2018ordinal} and Mathis et al. \cite{mathis2018deeplabcut} employing deep-learning frameworks to estimate body joint poses. Notably, OpenPose \cite{openpose2019} marked a significant milestone by introducing the first real-time multi-person system for body keypoints estimation, setting a benchmark for future research in human pose estimation. Additionally, Zhang et al. \cite{zhang2021deep} delved deeper into the extraction of human keypoints in natural settings without human labels, utilizing 3D point clouds.

Animal pose estimation, drawing inspiration from human pose estimation techniques, has advanced significantly in assessing various health traits and behaviours in animals. However, the field encounters challenges due to the scarcity of fully annotated datasets, hindering the effective application of deep learning methodologies.

Existing animal datasets, including those referenced in \cite{yu2021ap,russello2022t}, predominantly consist of 2D images or sequences without ground truth annotations for joint positions. While custom animal datasets like AwA for quadrupeds \cite{banik2021novel} exist, they heavily rely on manual annotation, limiting their suitability for autonomous training of deep learning models.

Due to the scarcity of datasets in the animal field, researchers have explored various strategies to augment data input. One approach involves adapting learning models from other domains, such as humans or different animal species, and fine-tuning them \cite{sanakoyeu2020transferring,mathis2021pretraining,pereira2020sleap,cao2019cross}. Other methods employ synthetic data augmentation processes \cite{mu2020learning,del2017behavior,del2015articulated,li2021synthetic,zuffi2019three}.

While some of these approaches extract 3D keypoints, they may not precisely correspond to actual joints \cite{badger20203d}, or the resulting 3D models may lack an evaluation of mesh quality \cite{yao2019monet,li2021synthetic}. Difficulties also arise in accurately assigning keypoint locations, necessitating interpretable relationships between the model and the actual animal shape \cite{zuffi2019three}. Furthermore, although visually appealing 3D shapes may be generated, they may not be geometrically consistent with the animal's structure \cite{yao2022lassie}.

Recent investigations into synthetic animated training data for 3D pose estimation of animals draw inspiration from human pose estimation methodologies~\cite{zoobuilder2020}. This process involves retraining established networks like OpenPose and Pose3D using simulated RGB and joint pose data generated under controlled conditions. The primary challenge is aligning the distribution of synthetic training data with that of real-world data collected from wildlife. While successful in addressing the difficulties of obtaining relevant anatomical keypoints on animal joints, this approach necessitates the conversion of RGB images into realistic environments. However, there have been no efforts to verify the differences between synthetic and real-world keypoint locations utilising 3D data for animal assessment.

Similarly, recent endeavours in extracting keypoints in 3D space, such as those proposed by other researchers \cite{yang2018semantic}, involve applying methods like heat kernel and geodesic distance. These keypoints are utilised for body segmentation, albeit positioned on the surface rather than within the joints. Alternatively, another work by \cite{Falque2022} suggests using multi-depth-camera systems and PointNet++ for extracting semantic keypoints. However, the keypoints in their dataset are manually annotated and primarily located on the surface rather than internally within the body.

Our work seeks to quantify the robustness of estimating joint locations, informed by a synthetic dataset containing a variety of shapes and animal poses. We do so by examining joint data on real cattle while they are in motion, as well as establishing a relationship between joint data and hip height inspired by \cite{ali1984relationship}. 

\section{Methodology}
\label{sec:methodology}

\begin{figure*}
  \centering
  \includegraphics[width=0.9\linewidth]{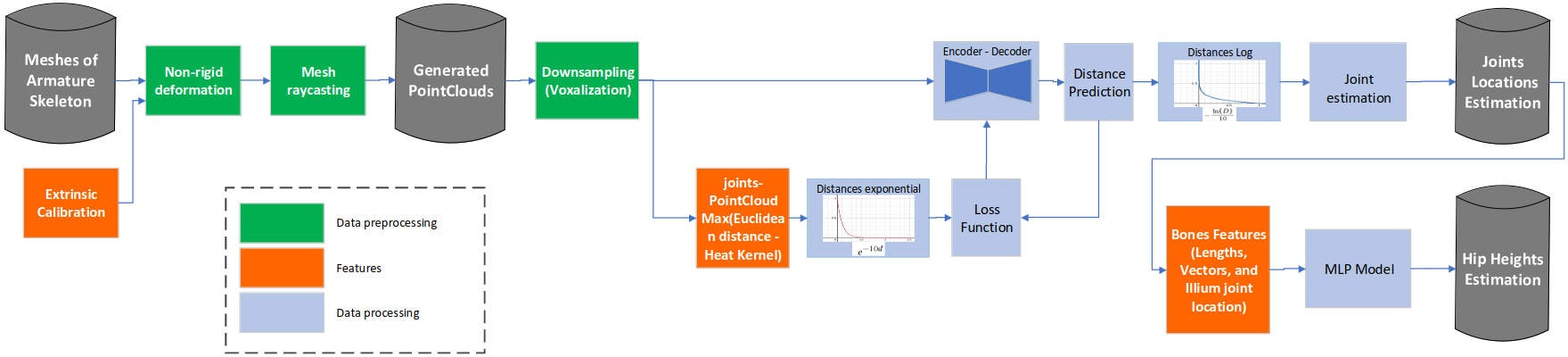}
  \caption{Method overview: from the simulated model, the armature undergoes rigid scaling and meshes a non-rigid deformation. Through raycasting over a number of cameras, several point clouds are generated and merged. At inference time, the merged point cloud is passed into an encoder-decoder architecture (Pointnet++~\protect\cite{qi2017pointnet++}) to extract the keypoints. During training, the dataset uses keypoints from the armature and the distances on the manifold are pre-computed. The encoder-decoder inputs are $\numberOfPoints\times3$ points, and the outputs are the $\numberOfPoints\times13$ distances to the $13$ joints keypoints.}
  \label{fig:systemdiagram}
\end{figure*}

\begin{figure}
  \centering
  \includegraphics[width=0.75\linewidth]{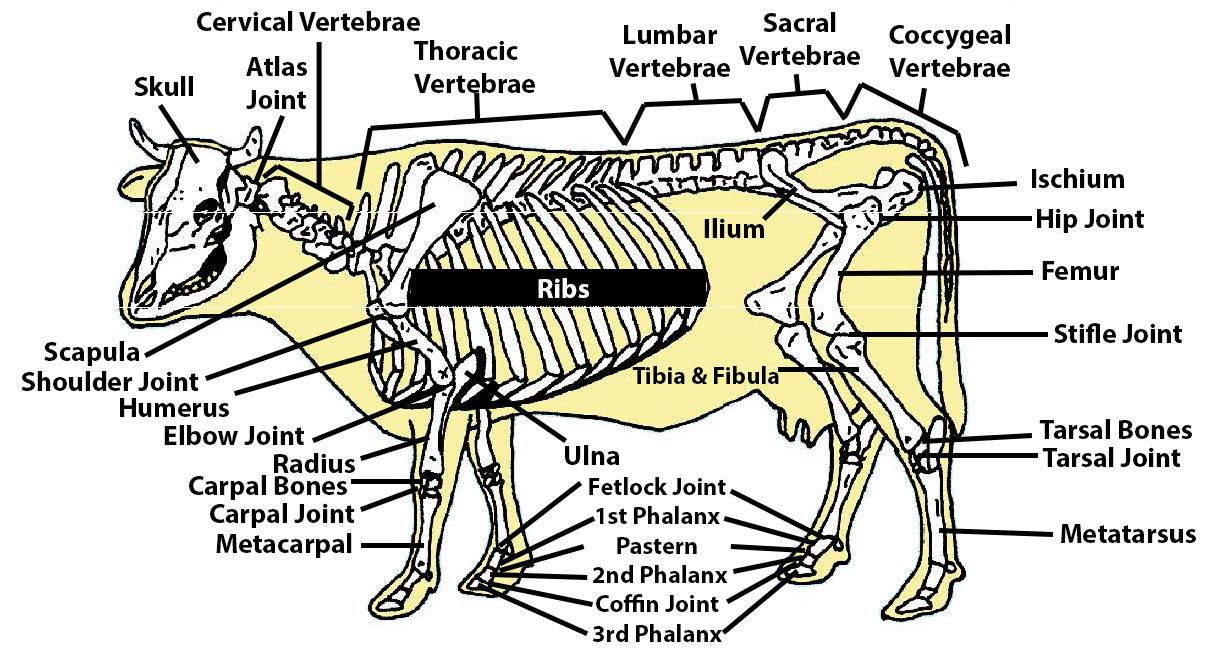}
  \includegraphics[width=0.75\linewidth]{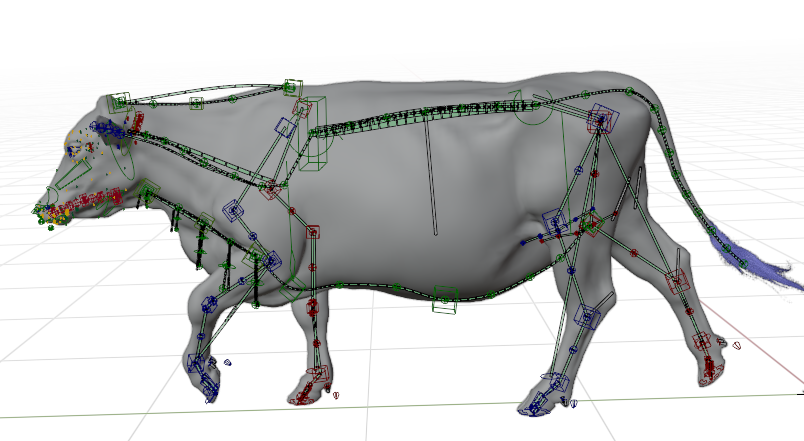}
  \caption{Top: Cattle skeleton from~\protect\cite{ukyBeefCattle} containing information of all the joints and bones. Bottom: The annotated model used in this work containing rigging and joints indicated by blue squares}
  \label{fig:Cattle_skeleton}
\end{figure}

 Given that joints are not inherently situated on the surface, their locations must be inferred from curvature on the surface via a deep-learning model. The system architecture, presented in Figure~\ref{fig:systemdiagram}, comprises four principal modules: point cloud creation, data preprocessing, a deep-learning model (PointNet++) tasked with estimating the nearest point cloud to each joint, and joint estimation. We also include an MLP model utilising joint information for hip height estimation. Apart from the annotated joints listed in \cite{okour2022sim2real}, which include Carpal joint left, Elbow joint left, Carpal joint right, Elbow joint right, Tarsal joint left, Stifle joint left, Hip joint left, Tarsal joint right, Stifle joint right, Hip joint right, and Front spine, a new keypoint positioned at the tip of the cattle's hip, termed the Illium joint, has been introduced. This is depicted in Figure~\ref{fig:Cattle_skeleton}. This new keypoint is located on the tip of the cattle's hip. This results in a total of twelve keypoints per animal, as illustrated in Figure~\ref{fig:Annotated model}, which are subject to prediction.

The PointNet model takes the point cloud location as input. As the joints lie outside the mesh surface, this necessitates the estimation of their positions from distances inferred by the deep learning model via multilateration. This method utilises Euclidean distance to determine the requisite points.

Small Euclidean distances between annotated joints within a point cloud can introduce significant errors during the PointNet estimation. In such cases, the model may struggle to distinguish between closely spaced joints, leading to inaccurate estimations. To address this, geodesic distances are computed as an additional input, incorporating information on the distance over the surface in the learning framework.

Data processing involves two primary steps. Firstly, geodesic distances are obtained, representing the distance on the manifold from each joint utilising the heat kernel method. Simultaneously, Euclidean distances from each joint to the generated point cloud are computed. Secondly, the maximum value between these two distance metrics is determined. The loss function is then computed based on the logarithmic value of the maximum distances carried by each point cloud, relative to a known mesh.

\begin{figure}
    \centering
    \includegraphics[width=0.5\linewidth]{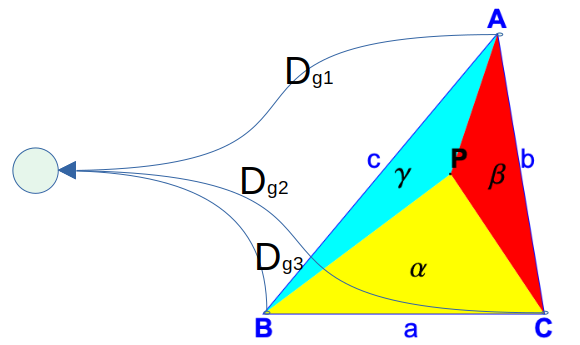}
    \caption{Barycetnric diagram of $\alpha$, $\beta$, and $\gamma$. Where $D_{g1}$, $D_{g2}$, and $D_{g3}$ are the heat kernel distances to a point in space.}
    \label{fig:barycentric}
\end{figure}

To compute the geodesic distances, we first downsample the generated point clouds and then utilise the heat kernel method~\cite{Crane:2017:HMD} to precompute the distances on the manifold $\distanceGaussian_g$. This computation leverages the tufted Laplacian~\cite{Sharp:2020:LNT} derived from the mesh structure. Subsequently, we employ a barycentric calculation, as depicted in Equation \ref{eq:barycentric_coords}, to determine the geodesic distance of each point cloud to the face vertices as shown in figure~\ref{fig:barycentric}. The output is used to find the final heat kernel distance $\distanceGaussian$ as illustrated in equation \ref{eq:barycentric_distance}.

\begin{equation}
\label{eq:barycentric_coords}
\begin{aligned}
\alpha &= \frac{{(\vec{AC} \cdot \vec{AC}) (\vec{AB} \cdot \vec{AP}) - (\vec{AB} \cdot \vec{AC}) (\vec{AC} \cdot \vec{AP})}}{{(\vec{AB} \cdot \vec{AB}) (\vec{AC} \cdot \vec{AC}) - (\vec{AB} \cdot \vec{AC})^2}} \\
\beta &= \frac{{(\vec{AB} \cdot \vec{AB}) (\vec{AC} \cdot \vec{AP}) - (\vec{AB} \cdot \vec{AC}) (\vec{AB} \cdot \vec{AP})}}{{(\vec{AB} \cdot \vec{AB}) (\vec{AC} \cdot \vec{AC}) - (\vec{AB} \cdot \vec{AC})^2}} \\
\gamma &= 1 - \alpha - \beta \\
B &= \begin{bmatrix}
    \alpha \\
    \beta \\
    \gamma
\end{bmatrix}
\end{aligned}
\end{equation}

\begin{equation}
\label{eq:barycentric_distance}
    D = \sum (B \times {\begin{bmatrix}
        D_{g1} \\
        D_{g2} \\
        D_{g3} 
    \end{bmatrix}})
\end{equation}

\begin{figure}
  \centering
  \includegraphics[width=0.49\linewidth]{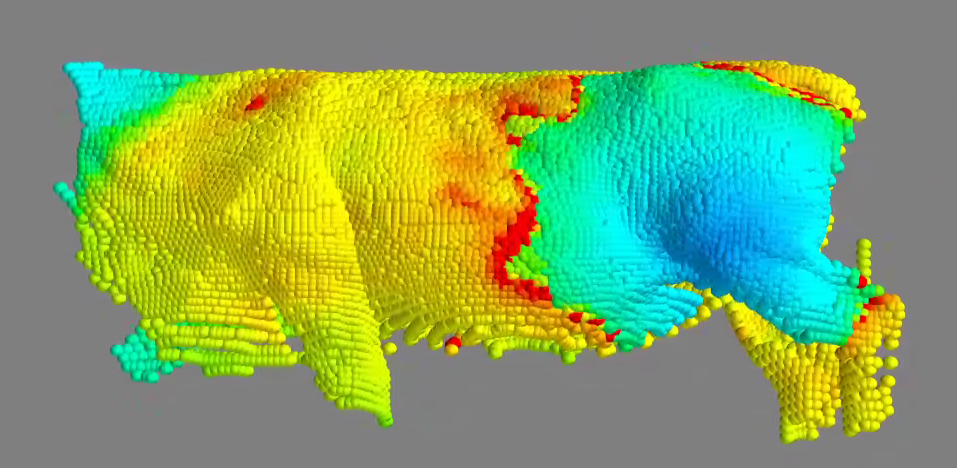}
  \includegraphics[width=0.49\linewidth]{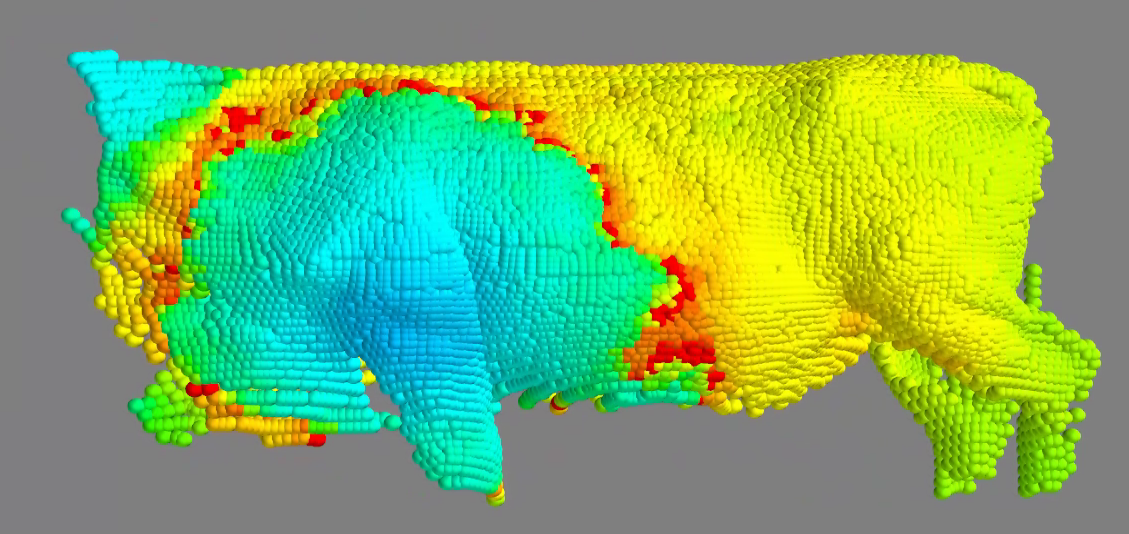}

  \caption{Predicted distance on the manifold. Points coloured in blue represent the nearest points to a joint. Left the rear leg and right the front leg are being evaluated.}
  \label{fig:distance_estimation}
\end{figure}

To accomplish this, during the creation of point clouds from the simulated model, we extract the hit face number for each point in the point cloud generated by ray casting. Subsequently, we consider the maximum distance ($D_{max}$) between the heat kernel distance on the manifold $\distanceGaussian$ and the Euclidean distance from a joint to each point in the point clouds, to be applied in the loss in the encoder-decoder network. We proposed to take the maximum distances to achieve a more accurate estimation of joint positions in cases where the joints are very close together, such as during walking. The logarithmic transformation is then applied to the maximum values to increase the penalty in the loss function as illustrated in equation \ref{eq:logarithmic_trans}. The effect of this transformation on one of the instances can be shown in Figure~\ref{fig:distance_estimation}

\begin{equation}
\label{eq:logarithmic_trans}
    L = -ln(D_{max})/10
\end{equation}

The encoder-decoder network takes as input the point cloud (i.e., a matrix $\shape$ of size $n \times 3$) and outputs a matrix $\hat{\distanceGaussian}$ of size $n \times m$, where $m$ represents the number of joints to be predicted.

\begin{figure}
  \centering
  \begin{subfigure}[b]{0.55\textwidth}
    \includegraphics[width=0.8\linewidth]{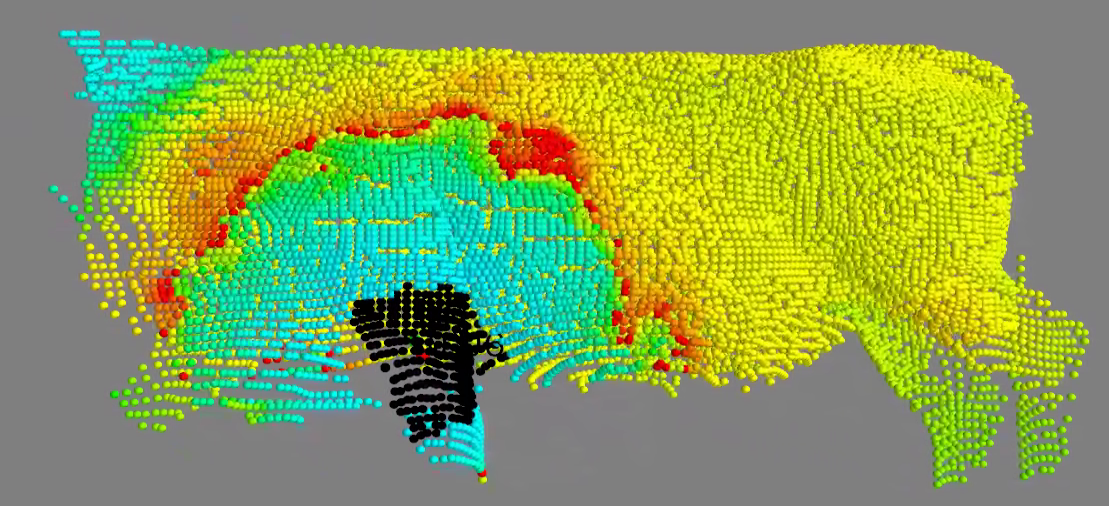}
    \caption{}
    \end{subfigure}
  \begin{subfigure}[b]{0.55\textwidth}  
    \includegraphics[width=0.8\linewidth]{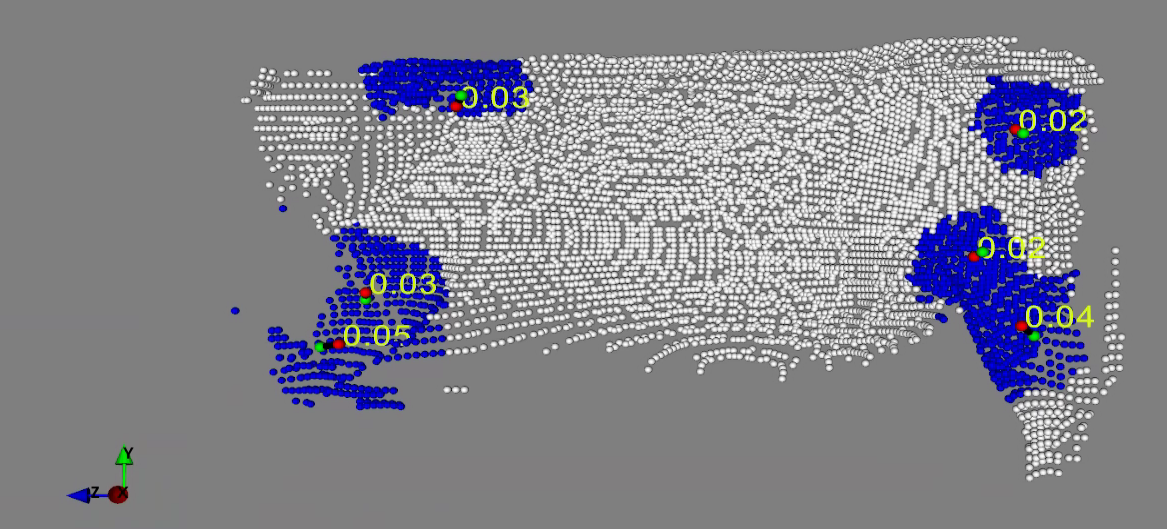}
    \caption{}
    \end{subfigure}
  \caption{(a): The blue points represent the complete set of the lowest estimated $D_{max}$ values. The area of interest, shown in black, highlights a subset of these points with the lowest estimated $D_{max}$. This output corresponds to the front left joint.  (b) Represents joints prediction of an instance (red spheres are the ground truth and green are the estimated joints) with the error between both in meters}
  \label{fig:predicted_joints}
\end{figure}

The final step in predicting joints involves utilising the distance predictions from the PointNet++ model to estimate the positions of the joints. In line with prior research~\cite{tarrio2011weighted}, we propose employing the multilateration technique to enhance joint estimation accuracy as in equation \ref{eqn: multilateration}. Specifically, we designate the first point within the area of interest of the point cloud as the anchor and apply the least squares method to determine the joint positions. This area of interest constitutes a subset of points with the lowest estimated $D_{max}$, as illustrated in Figure \ref{fig:predicted_joints}. Formulations \ref{eq:close_form} and \ref{eq:joint_prediction} allow predicting joints' position outside of the point cloud in the case where the underlying shape is convex, where $d_n$ is the $D_{max}$. A sample of the joint detection using multilateration of the nearest point cloud group to a joint is displayed in Figure~\ref{fig:predicted_joints}.

\begin{equation}
    \varepsilon = \sum_{i=1}^{N} (\sqrt{(x_{i}-x)^2+(y_{i}-y)^2+(z_{i}-z)^2)} - \Tilde{d_i})^2
    \label{eqn: multilateration}
\end{equation}
\begin{equation}
    \begin{bmatrix}
        2x_{2} & 2y_{2} & 2z_{2}\\
        \vdots & \vdots & \vdots \\
        2x_{n} & 2y_{n} & 2z_{n}
    \end{bmatrix}
    \begin{bmatrix}
        {x}\\
        {y} \\
        {z}
    \end{bmatrix}
    = \begin{bmatrix}
        x_{2}^2 + y_{2}^2 + z_{2}^2 - d_2^2 + d_1^2\\
         \vdots \\
        x_{n}^2 + y_{n}^2 + z_{n}^2 - d_n^2 + d_1^2
    \end{bmatrix}
    \label{eq:close_form}
\end{equation}

\begin{equation}
    \hat{x} = (H^TH)^{-1}H^T\Tilde{b}
    \label{eq:joint_prediction}
\end{equation}

Where H = $\begin{bmatrix}
            2x_{2} + 2y_{2} + 2z_{2}\\
         \vdots \\
        2x_{n} + 2y_{n} + 2z_{n}
        \end{bmatrix}$,
        
        $\Tilde{b} = \begin{bmatrix}
        x_{2}^2 + y_{2}^2 + z_{2}^2 - d_2^2 + d_1^2\\
         \vdots \\
        x_{n}^2 + y_{n}^2 + z_{n}^2 - d_2^2 + d_1^2
    \end{bmatrix}$ ,
    and $\hat{x} = \begin{bmatrix}
        {x}\\
        {y} \\
        {z}
    \end{bmatrix}$

We thereafter employ a Multilayer Perceptron (MLP) model, consisting of 3 hidden layers (with 9, 7, and 5 neurons for each layer) with ReLU activation functions, to estimate the hip height. The model is trained using the Adam optimiser on a dataset comprising 175 instances of real animals, applying the leave-one-out technique to ensure fair evaluation. The features used for this prediction contain the coordinates of the keypoint related to the hip of the animal and information related to the backbones (i.e., their length and the vector from joint to joint). The backbones used in this module include Femur, Tibia, and Fibula. While the hip keypoint is the joint between the Ilium and Lumbar Vertebrae bones, as indicated in figure \ref{fig:Cattle_skeleton}.

In summary, we employ the heat kernel method \cite{Crane:2017:HMD} to compute the maximum value between the distance on the manifold $\distanceGaussian$, obtained via the tufted Laplacian \cite{Sharp:2020:LNT} on the mesh, and the euclidean distance from each joint to the point clouds. Subsequently, we utilise a barycentric calculation to derive the geodesic distance of each point cloud to the vertices of each triangular face of the mesh. This distance on the manifold is then treated as a feature and learned using an encoder-decoder network. The inputs of the encoder-decoder consist of the point cloud size, represented by a matrix $\shape$ of size $n \times 3$, while the outputs correspond to the number of joints to be predicted, represented by a matrix $\hat{\distanceGaussian}$ of size $n \times m$.

\section{Experiments}
\begin{figure}
  \centering
  \includegraphics[width=\linewidth]{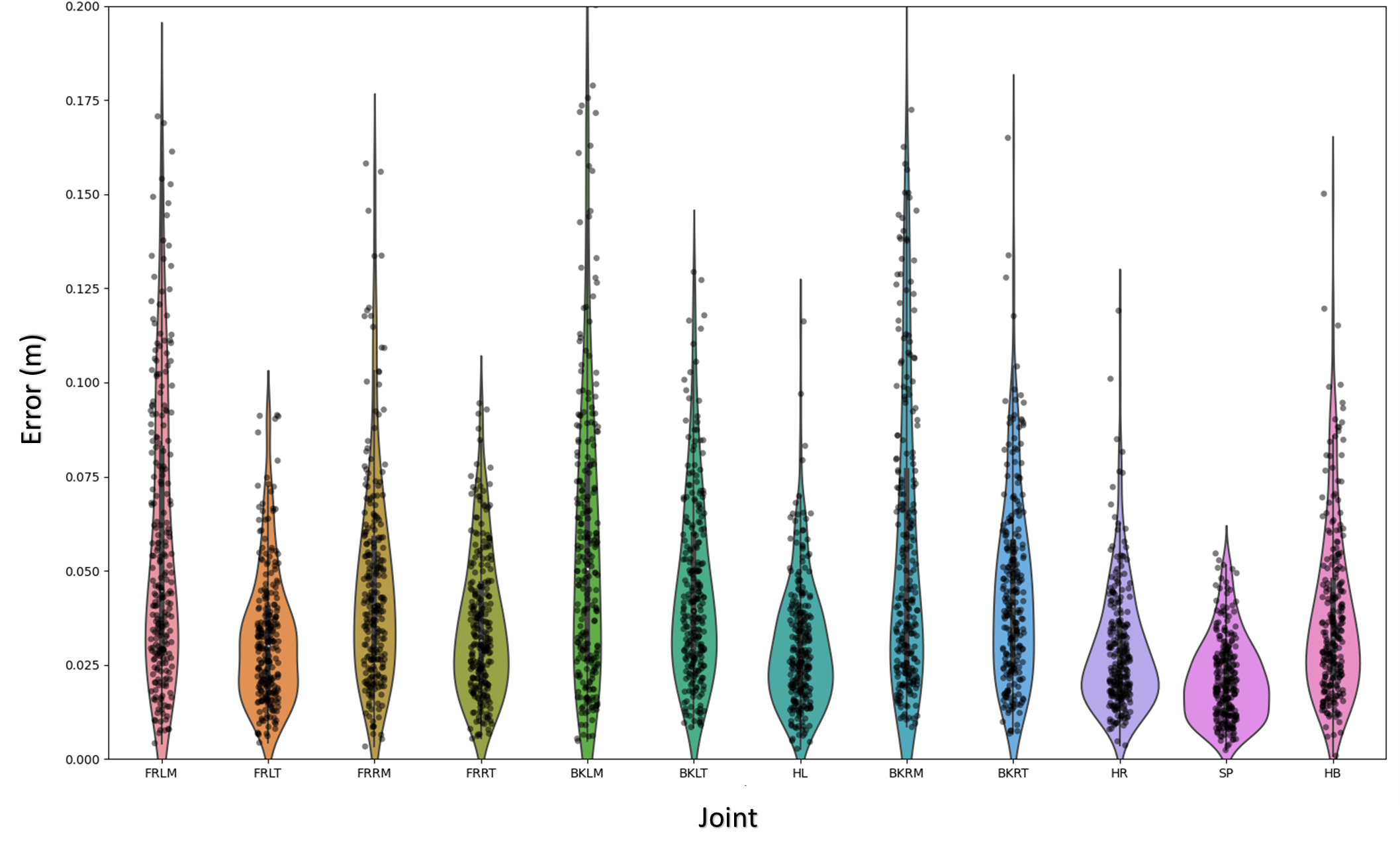}
  \caption{Joints estimation error of synthetic data set. Joint names starting from left to right: Carpal joint left, Elbow joint left, Carpal joint right, Elbow joint right, Tarsal joint left, Stifle joint left, Hip joint left, Tarsal joint right, Stifle joint right, Hip joint right, Front spine, and Illium joint}
  \label{fig:joints_error}
\end{figure}

Utilising the annotated dataset derived from \cite{okour2022sim2real} and employing the joint prediction methodology detailed in Section~\ref{sec:methodology}, we calculate the joints estimation error based on a simulated test dataset, as illustrated in Figure \ref{fig:joints_error}. The analysis reveals a mean error in joints estimation of 0.03 and a standard deviation of 0.01.

\begin{figure}
  \centering
  \includegraphics[width=\linewidth]{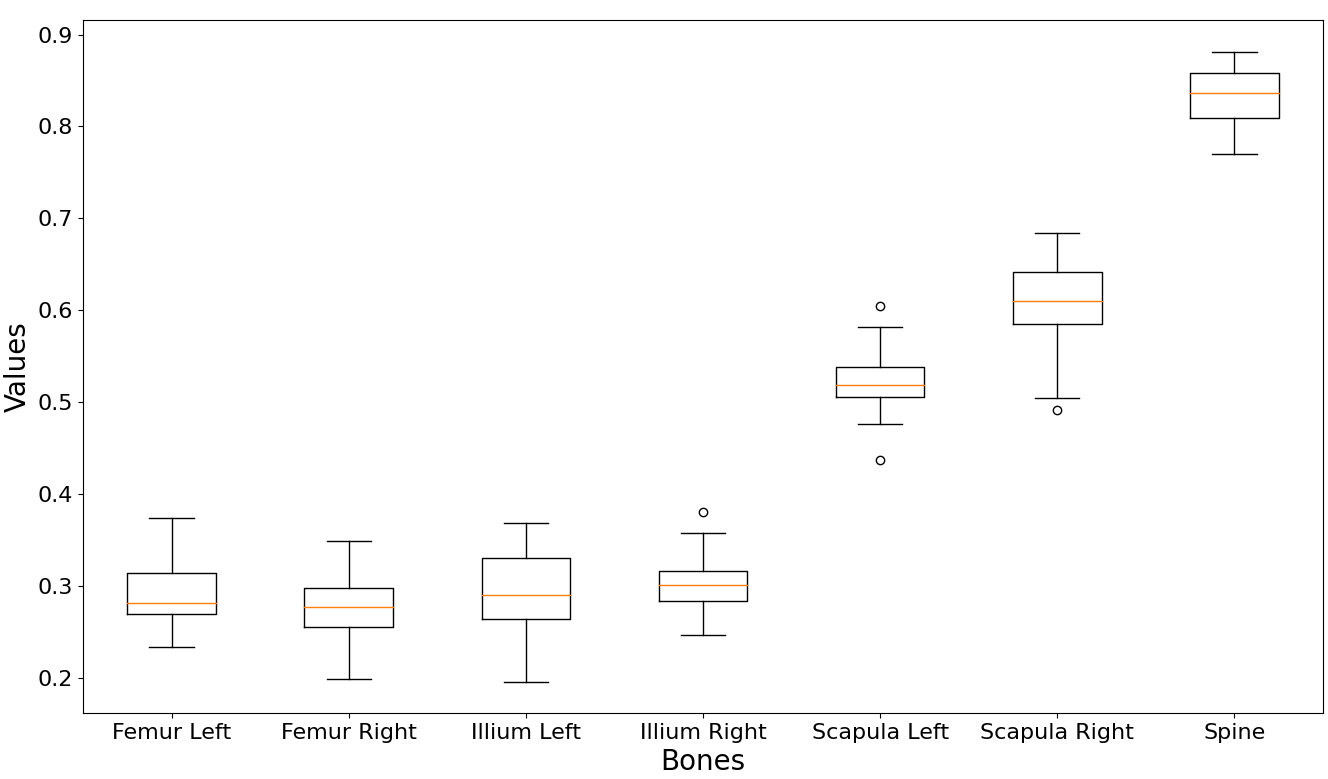}
  \caption{Bone length estimation on a walking animal. This estimation was performed over eighteen consecutive frames while the animal walked through the race. For each bone length, we denote the mean, standard deviation and minimum/maximum estimate on the box plot. The scapula and spine are the longest bone structures.}
  \label{fig:real_bones_length}
\end{figure}

We assess the \emph{sim2real} gap, resulting from training the network on a simulated dataset, via the consistency of joint predictions on a real animal walking through the race. We analyse the variation in bone length estimations and distance between joints across eighteen consecutive frames (instances); the findings of this evaluation are detailed in Figure~\ref{fig:real_bones_length}, shedding light on the stability and reliability of the model's predictions of joint location in real-world scenarios.

\begin{figure}
  \centering
  \includegraphics[width=0.7\linewidth]{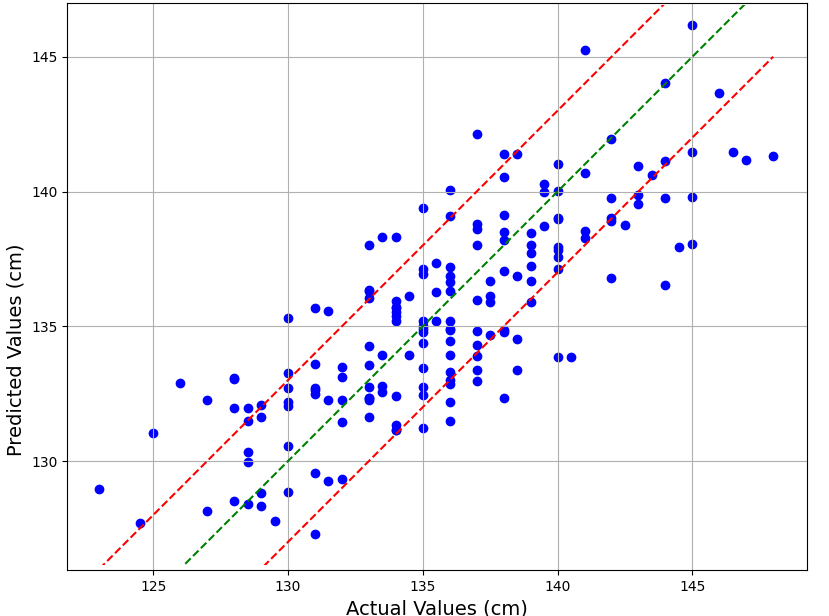}
  \caption{Predicted vs actual values of hip height estimation for a dataset comprising 175 instances of real animals, using the leave-one-out technique. The $R^2$ score is 0.64, and the Root Mean Square Error (RMSE) is 2.97. The green dashed line represents the ideal output, while the red dashed lines indicate a margin of $\pm$3.}
  \label{fig:hip_height_error}
\end{figure}

A further quantitative evaluation is conducted by estimating the hip height of 175 real animals where the cattle were subsequently restrained to manually measure hip height. This evaluation is depicted in Figure~\ref{fig:hip_height_error}. As detailed in Section~\ref{sec:methodology}, the features employed for this prediction encompass the length of the back bones, the geometric vectors associated with these bones, and the highest point among the estimated nearest points of the hip keypoint. 

\section{Discussion}
The results obtained from the joint estimation error analysis reveal a mean error in joint estimation of 0.026 and a standard deviation of 0.012, with the majority of errors falling within the range of 100 mm. This indicates a certain level of accuracy in the joint prediction methodology utilised, although there is room for improvement, particularly in reducing the maximum error.

Moving on to the quantitative evaluation of the \emph{sim2real} gap through hip height estimation, we observe a coefficient of determination ($R^2$) of 0.64 and a root mean square error (RMSE) of 2.97. Despite the absence of certain segments within the real animal testing dataset, these metrics suggest a moderate level of predictive capability, implying that the trained network can reasonably estimate hip height based on the features employed.

Figure~\ref{fig:real_bones_length} illustrates the model's predictions of bone lengths for 18 consecutive walking cattle, offering insights into the accuracy and reliability of these predictions. Our findings reveal a smaller difference between the minimum and maximum values, indicating minimal variation in bone lengths during walking motion and suggesting the model's accuracy in predicting such changes. This consistency across diverse bones, including the left and right femur, ilium, scapula, and spine, underscores the precision of the model's estimations. These results contribute to the understanding of how predictive models can effectively estimate bone lengths in dynamic scenarios, such as walking, in cattle.

Overall, these results demonstrate the effectiveness of the employed methodologies in estimating joint positions, hip height, and bone lengths, with some limitations and opportunities for further refinement.

\section{Conclusion}
This study presents a novel method for predicting joints using synthetic auto-annotated datasets. By estimating keypoints outside the mesh and utilising created point clouds, we address the challenge of bridging the gap between training deep models with simulated datasets and real-world data for this specific application.

Although our findings demonstrate promising results in joint prediction, further exploration is warranted to evaluate the potential applicability of this approach to other domains, such as health-related metrics. Additionally, future research could explore the development of anatomically accurate models for muscle and fat layers attached to the joints. Creating such models is a complex task typically undertaken by only a select few highly skilled 3D artists. Moreover, making these models parametric to encompass various body conditions for extensive data augmentation would introduce additional challenges to the process.

\section*{Acknowledgments}
This work was partly supported by an Australian Government Research Training Program (RTP) Scholarship, Food Agility CRC top-up scholarship, the University of Technology Sydney and Meat and Livestock Australia under grant number B.GBP.0051. The use of animals and the procedures performed in this study were approved by the University of New England (UNE) Animal Ethics Committee (Approval number: ARA21-070).

\bibliographystyle{IEEEtran}
\bibliography{iros2024_sim2real}

\end{document}